\documentclass[letterpaper, 10 pt, conference]{ieeeconf}  % Comment this line out if you need a4paper

\IEEEoverridecommandlockouts                              % This command is only needed if 
\usepackage{amsmath} % assumes amsmath package installed
\usepackage{amssymb}  % assumes amsmath package installed
\usepackage{amsfonts}
\usepackage{dsfont}
\usepackage{esvect}
\usepackage{graphicx}
\usepackage{empheq}
\usepackage{siunitx}
\usepackage{bbold}
\usepackage{textcomp}
\usepackage{array,multirow,makecell}
\usepackage{hyperref}

\usepackage{tikz}
\usetikzlibrary{shapes,arrows}
\usetikzlibrary{arrows.meta}
\usetikzlibrary{angles,quotes}
\usetikzlibrary{3d}
\usepackage{amsmath,bm,times}
\usepackage{verbatim}

\newcommand\real{{\mathbb R}}
\newcommand{\skewsym}[1]{\left[#1\right]_{\times}}

\DeclareMathOperator{\diag}{diag}

\newtheorem{remm}{Remark}

\newcommand{\rotateRPY}[4][0/0/0]% point to be saved to \savedxyz, roll, pitch, yaw
{   \pgfmathsetmacro{\rollangle}{#2}
    \pgfmathsetmacro{\pitchangle}{#3}
    \pgfmathsetmacro{\yawangle}{#4}

    \pgfmathsetmacro{\newxx}{cos(\yawangle)*cos(\pitchangle)}% a
    \pgfmathsetmacro{\newxy}{sin(\yawangle)*cos(\pitchangle)}% d
    \pgfmathsetmacro{\newxz}{-sin(\pitchangle)}% g
    \path (\newxx,\newxy,\newxz);
    \pgfgetlastxy{\nxx}{\nxy};

    \pgfmathsetmacro{\newyx}{cos(\yawangle)*sin(\pitchangle)*sin(\rollangle)-sin(\yawangle)*cos(\rollangle)}% b
    \pgfmathsetmacro{\newyy}{sin(\yawangle)*sin(\pitchangle)*sin(\rollangle)+ cos(\yawangle)*cos(\rollangle)}% e
    \pgfmathsetmacro{\newyz}{cos(\pitchangle)*sin(\rollangle)}% h
    \path (\newyx,\newyy,\newyz);
    \pgfgetlastxy{\nyx}{\nyy};

    \pgfmathsetmacro{\newzx}{cos(\yawangle)*sin(\pitchangle)*cos(\rollangle)+ sin(\yawangle)*sin(\rollangle)}
    \pgfmathsetmacro{\newzy}{sin(\yawangle)*sin(\pitchangle)*cos(\rollangle)-cos(\yawangle)*sin(\rollangle)}
    \pgfmathsetmacro{\newzz}{cos(\pitchangle)*cos(\rollangle)}
    \path (\newzx,\newzy,\newzz);
    \pgfgetlastxy{\nzx}{\nzy};

    \foreach \x/\y/\z in {#1}
    {   \pgfmathsetmacro{\transformedx}{\x*\newxx+\y*\newyx+\z*\newzx}
        \pgfmathsetmacro{\transformedy}{\x*\newxy+\y*\newyy+\z*\newzy}
        \pgfmathsetmacro{\transformedz}{\x*\newxz+\y*\newyz+\z*\newzz}

    }
}

\tikzset{RPY/.style={x={(\nxx,\nxy)},y={(\nyx,\nyy)},z={(\nzx,\nzy)}}}
\title{\LARGE \bf
Modelling and hovering stabilisation of a free-rotating wing UAV}

\author{Florian Sansou$^{1}$, Gautier Hattenberger$^{1}$, Luca Zaccarian$^{2}$, Fabrice Demourant$^{3,1}$, Thomas Loquen$^{3,1}$
\thanks{Research supported in part by Fédération ENAC ISAE-SUPAERO ONERA, Université de Toulouse, France and by Occitanie region.}
\thanks{$^{1}$~Fédération ENAC ISAE-SUPAERO ONERA, Université de Toulouse, France, (e-mail: firstname.lastname@enac.fr)}
\thanks{$^{2}$~Department of Industrial Engineering, University of Trento, Italy, and LAAS-CNRS, Universit\'e de Toulouse, CNRS, Toulouse, France, (e-mail: zaccarian@laas.fr).}
\thanks{$^{3}$~DTIS, ONERA,  Université de Toulouse, 31000, Toulouse, France (e-mail: firstname.lastname@onera.fr)}
}

\begin{document}

\maketitle
\thispagestyle{empty}
\pagestyle{empty}

%%%%%%%%%%%%%%%%%%%%%%%%%%%%%%%%%%%%%%%%%%%%%%%%%%%%%%%%%%%%%%%%%%%%%%%%%%%%%%%%
\begin{abstract}
% \todo{desirable turbulence rejection
% properties remove ,  with desirable turbulence rejection properties}
We propose a multibody model of a freewing UAV. This model allows obtaining simulations of the UAV's behaviour and, in the future, to design a control law stabilising the entire flight envelope (hovering and forward flight). We also describe the realisation of a prototype and a comparison of possible methods for estimating the UAV's states. With this prototype, we report on experimental hovering flights with a non-linear incremental dynamic inversion controller to stabilise the wing and a proportional derivative controller for the fuselage stabilization.
\end{abstract}
% \listoftodos

%%%%%%%%%%%%%%%%%%%%%%%%%%%%%%%%%%%%%%%%%%%%%%%%%%%%%%%%%%%%%%%%%%%%%%%%%%%%%%%%
\section{Introduction}
Tail-sitter UAVs are very sensitive to turbulence, as they have a large vertical wing area when hovering. However, there are many advantages with this type of architecture, such as the possibility of transitioning to flying like an aeroplane thus having a long range. However, a problem arises when carrying a payload.
% In fact, the payload rotates during the transition, which is undesirable, e.g., when the payload is a camera. Similarly, all the sensors are subject to this rotation, making it impossible to measure the wind speed in certain configurations, because the Pitot probe is not parallel to the air stream.
% \todo{check}
% GH:
Since the body rotates during the transition, payloads such as cameras may end in an undesirable position or orientation. Similarly, other sensors are subject to this rotation, which can make them unusable like airspeed sensors when not aligned with air stream.

In this context, we designed an architecture that retain the properties of a convertible drone, but where we can mount sensors whose orientation can be kept constant throughout all the flight phases. The idea is to install the wing on a pivot on its pitch axis, giving it freedom of rotation by separating its movement from that of the fuselage. As the weights are on the fuselage, the wing's inertia is lower, allowing the wing to naturally settle in the direction of the wind. This idea dates back to a patent published by \cite{patentZuck}. 

Since then, a great deal of work has been produced using this architecture. Using a freewing has its advantages and disadvantages as discussed in \cite{Date1972ExperimentalIO}. The compromise is between gust attenuation and low-speed performance. The gust attenuation is obtained from a fast convergence speed of the pitch wing axis, which is proportional to the static stability of the wing. However, the elevons generate an out-of-balance force to stabilise the moment equation. In addition, \cite{Date1972ExperimentalIO} mentions the possibility of wing flutter in this configuration. 
As the motors are installed on the wing, they generate a permanent air flow over part of the wing, thus preventing stalling. In addition, the motors used for hovering also generate traction during forward flight, wherein it is necessary to switch off the motor that stabilises the fuselage. As a result, the drone carries very little unnecessary actuator weight in the high efficiency forward flight phases. Others architectures, such as coplanar UAVs generate large parasitic drag such as Delair DT46, VTOL version \cite{Delair_2024}.
% Idéalement, il faudrait une ref, ou au moins un exemple de 'coplanar' UAV

Key parameter in the design is the position of the wing pivot: by moving the wing pivot point, it is possible to vary the wing's natural modes \cite{doi:10.2514/6.1992-4194}.

An important part of the study of a UAV is in its modelling. The complexity of freewing UAVs arises from their multibody modelling, induced by the degree of freedom between the wing and the fuselage. Modelling based on the Newton-Euler equations, with a representation of the orientation using Euler angles, makes it possible to appreciate the complexity of such a system \cite{doi:10.2514/6.2005-1024, RoKammanJames}.
Other studies show the use of free links to reject disturbances, in particular by cutting the wing to allow the tips to freely change their incidence. Multi-body modelling is often carried out to represent the different parts of the wing \cite{Leylek2015UseOC}.

A long sequence of work has been proposed: modelling and simulation-based analysis of the vibration modes \cite{HavBerJoh}, hovering stabilisation of the UAV using a PI-Feed Forward (PIFF) control law \cite{doi:10.2514/6.2020-1263}. Finally, a work based on a multi-outer loop dynamic inversion control law was presented in \cite{doi:10.2514/6.2024-0726} with simulations and outdoor flights in windy conditions. 

Later work has studied tilt-wing UAVs, where the wing is not free to rotate about the pitch axis, but is controlled by an actuator \cite{McSwain2017GreasedL}. However, nontrivial couplings between the actuators (propeller and elevon) emerge. The interesting aspect is the use of a control law based on the INDI \cite{binz2019attitude}.

A final important task after modelling is identifying the coefficients, and a number of methods have been proposed based on computational fluid dynamics (CFD) \cite{akr2022DesignAA} or wind tunnel measurements  \cite{articleSimmonsMurphy, schutt2014fullscale}.

%Objectif
% \todo{add explanation of the contribution :This paper appears to be about the derivation of
% dynamics, and filter design of a freewing UAV using
% quaternions. Is designing new hardware your main
% contribution? From this point of view, the introduction
% should specifically be stated what the clear contribution
% of this paper is.}
The main contribution of this paper is a model of our freewing UAV architecture based on the Udwadia-Phohomsiri equations \cite{udwadia-phohomsiri, udwadia-schutte, udwadia-koganti}, which represent the dynamics of a multi-body UAV based on constraints expressed between the bodies. This method makes it possible to express the forces on each part independently and to obtain dynamical equations representing the coupling. In future work, this modelling will enable designing nonlinear controllers for the UAV. The complete model and simulation of the UAV are discussed in the Section \ref{sec:model}.
After obtaining a model, we determine the ideal location for the autopilot. Section \ref{sec:stateEst} discusses the state estimation of the UAV on the real model, obtained from a rotary encoder used to measure the angle between the wing and the fuselage. We use a high-gain filter to estimate the speed from the rotary encoder measurements, allowing us to estimate all the states of the model. Section \ref{sec:control} describes the control architecture chosen to stabilise the UAV (wing and fuselage) in a hovering configuration. Based on two decentralized control loops, we use Incremental Nonlinear Dynamic Inversion (INDI) to stabilize the wing and a PD feedback loop to stabilize the fuselage. Section \ref{sec:exp} presents the results of the experimental flights in a controlled environment using the Paparazzi UAV (Unmanned Aerial Vehicle) open-source drone hardware and software project. Paparazzi provides the guidance, navigation and stabilisation layer for a set of UAVs based on a modular architecture, where the user can choose the different codes executed on the drone.

\noindent
{\bf Notation}.
Given two vectors $x_1$ and $x_2$, we often denote their juxtaposition as $(x_1,x_2):=
[x_1^\top \; x_2^\top]^\top$. Given any vectors $u, v\in \real^3$, the skew-symmetric matrix $\skewsym{u}$ satisfies $\skewsym{u} v = u \times v$.  The symbol $I$ denotes the identity matrix of appropriate dimensions. $(G)^{\dag}$ denotes the left Moore–Penrose matrix pseudo-inverse of $G$.

\section{Design and modelling of Colibri UAV}\label{sec:model}
\begin{figure}[h]
\centering
    \includegraphics[width=1\columnwidth,angle=0,trim={0 0 0 0.5cm},clip]{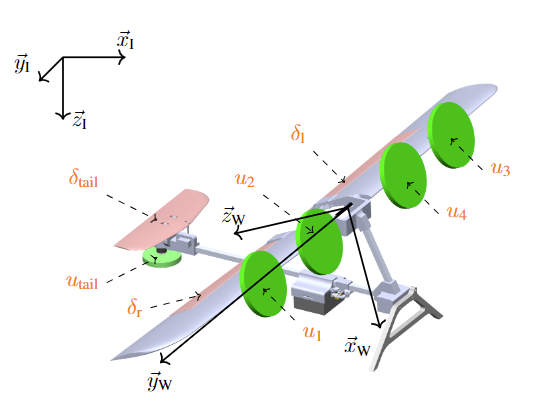}
    \caption{Inertial (I) and wing (W) reference frames and the Colibri architecture. }
    \label{fig:world_body}
\end{figure}

The Colibri drone is derived from a tail-sitter drone with a wing that generates lift during the forward flight. This wing has several actuators: four motors $u_{i}, ~i = 1,2,3,4$ and two elevons $\delta_{\text{l}}$ and $\delta_{\text{r}}$. We can define the control vector $u_{\text{W}}$ of the wing based on Figure~\ref{fig:world_body} as $u_{\text{W}} = [u_{1}~u_{2}~u_{3}~u_{4}~\delta_{\text{l}}~\delta_{\text{r}}]^\top$. A fuselage linked by a pivot is secured at the aerodynamic centre of the wing. This fuselage supports the autopilot, the battery, a motor and a tail to keep it horizontal. In Figure \ref{fig:world_body}, all the aerodynamic control surfaces are shown in pink and the propellers are shown in green.
There are three reference frames attach to the drone. (I) is a NED inertial reference frame (or world frame) linked to the earth's surface, (W) is a wing reference frame attached to the drone wing and (F) is a fuselage reference frame attached to the drone fuselage.
% Deux configurations sont étudiées pour la position des moteurs. 

% \begin{figure}[h]
% \centering
%     \includegraphics[width=1\columnwidth,angle=0]{picture/colibriH.jpg}
%     \caption{The Colibri convertible UAV, H version. }
%     \label{fig:colibri_h}
% \end{figure}

% L'architecture en H offre un actionnement important 

% \begin{figure}[h]
% \centering
%     \includegraphics[width=1\columnwidth,angle=0]{picture/ColibriV2.png}
%     \caption{The Colibri convertible UAV, motor on the wing. }
%     \label{fig:colibriv2}
% \end{figure}

% \todo{he reference frames could be introduced
% and detailed at the beginning of the section}

Some of the characteristic dimensions are shown in Table~\ref{tab:pars}. Note that the motors are positioned symmetrically on the wing, which means that the position can be described by focusing on one side. 
\begin{table}[ht]
  \centering
    \begin{tabular}{|l|c|c|}
      \hline
      \multicolumn{1}{|c|}{Parameter} & Value & Units  \\
      \hline
      $m_{\text{W}}$ (wing mass)  & 0.53& \SI{}{\kilogram} \\
      \hline
      $m_{\text{F}}$ (fuselage mass)  & 1.17& \SI{}{\kilogram} \\
      \hline
    %   $b$ (wingspan)  & 1.17 & \SI{}{\meter} \\
    %   \hline
    %   $c$ (aerodynamic cord)  & 0.150 & \SI{}{\meter} \\
    %   \hline
    %   $S$ (wing area) & 0.1537 & \SI{}{\square\meter}\\
    %   \hline
    %   $S_{\text{wet}}$ (wet area) & 0.0813 & \SI{}{\square\meter}\\
    %   \hline
    %   $S_{\text{p}}$ (propeller area) & 0.0182 & \SI{}{\square\meter}\\
    %   \hline
      $J_{\text{W}}=diag(J_{x}^{\text{W}}, J_{y}^{\text{W}}, J_{z}^{\text{W}})$ & \!\! $\diag(0.1677,0.0052,0.1634)$\!\! & \SI{}{\kilogram\square\meter}\\
      \hline
      $J_{\text{F}}=diag(J_{x}^{\text{F}}, J_{y}^{\text{F}}, J_{z}^{\text{F}})$ & \!\! $\diag(0.0191,0.0161,0.0343)$\!\! & \SI{}{\kilogram\square\meter}\\
      \hline
      $k_{\text{f}}$ (propeller thrust coeff.) & 1.7800e-8 & \SI{}{\kilogram\meter}\\
      \hline
    %   $k_{\text{m}}$(propeller torque coeff.) & 2.1065e-10 & \SI{}{\kilogram\square\meter}\\
    %   \hline
    %   $p_{z}^{int}$ (propeller $z$ location) & 0.037 & \SI{}{\meter}\\
    %   \hline
    %   $p_{y}^{int}$ (propeller $y$ location) & 0.145 & \SI{}{\meter}\\
    %   \hline
    %   $p_{z}^{ext}$ (propeller $z$ location) & 0.028 & \SI{}{\meter}\\
    %   \hline
    %   $p_{y}^{ext}$ (propeller $y$ location) & 0.325 & \SI{}{\meter}\\
    %   \hline
    %   $\xi_{\text{f}}$ (elevons lift coeff.) & 0.2 & --\\
    %   \hline
    %   $\xi_{\text{m}}$ (elevons torque coeff.) & 1.4 & --\\
    %   \hline
    %   $\rho$ (air density) & 1.225 & \SI{}{\kilogram\per\cubic\meter}\\
    %   \hline
    %   $C_{\text{d0}}$ (drag coeff.) & 0.1644 & --\\
    %   \hline
    %   $C_{\text{l0}}$ (lift coeff.) & 5.4001 & --\\
    %   \hline
       $d_{\text{M}O_{\text{W}}}$  & $[0.383,0,-0.167]^\top$ & \SI{}{\meter}\\
      \hline
       $d_{\text{G}O_{\text{W}}}$  & $[0.052,0,-0.171]^\top$ & \SI{}{\meter}\\
      \hline
    \end{tabular}
    \caption{\label{tab:pars} Numerical parameters of the Colibri model.}
\end{table}

% \todo{There is no derived process or explanation provided for
% the M matrix, A, Q matrix, etc. In Equation 1, both Q and B
% are matrices and do not include the location x. Where are
% the state variables located?}
The modelling is based on the results of \cite[Section 2.15]{udwadia-phohomsiri}. The algorithm for computing matrices $M$, $A$, $Q$ and $B$ is in \cite{udwadia-schutte}, which provides us the equations of motion of a constrained multibody system: 
\begin{align}
\label{eq:udwadia}
    \ddot{x} = \hat{M}^{\dag} \begin{bmatrix} Q \\ B \end{bmatrix}  = \begin{bmatrix} (I - A^{\dag}A)M \\ A \end{bmatrix}^{\dag} \begin{bmatrix} Q \\ B \end{bmatrix}
\end{align}
whose expression is valid as long as $\hat{M}$ has full rank and where $A$, $M$, $Q$ and $B$ are described next.\\
\begin{figure}[h]
\centering
    \includegraphics[width=1\columnwidth,angle=0,trim={0 0 0 0.5cm},clip]{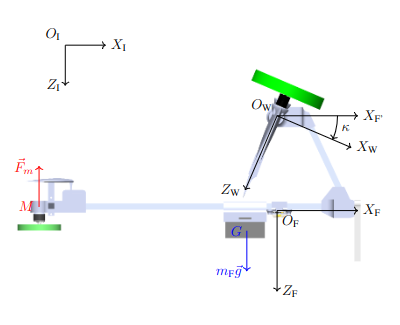}
    \caption{Inertial (I), fuselage (F) and wing (W) reference frames and forces acting on the Colibri UAV.}
    \label{fig:colibri_frame_side}
\end{figure}
We will use quaternions $q = \left [ \eta ~ \epsilon^\top \right]^\top  \in {\mathbb S}^3:=\{ q\in \real^4: |q| = 1\}$ to represent the orientations of the two bodies.  The ensuing rotation matrix $R(q) \in SO(3): = \{R\in \real^{3\times 3}: \; 
 R^\top R = I, \det (R)=1\}$ is uniquely defined as 
$R(q) := I +2\eta \skewsym{\epsilon} + 2\skewsym{\epsilon}^{2} = [R_{1}~R_{2}~R_{3}]$. \\
According to Figure \ref{fig:world_body} and \ref{fig:colibri_frame_side}, define the vectors $p_{\text{F}} = \overrightarrow{O_{\text{I}} O_{\text{F}}} $, $p_{\text{W}} = \overrightarrow{O_{\text{I}} O_{\text{W}}} $, $d_{\text{FW}} = \overrightarrow{O_{\text{F}} O_{\text{W}}} $ satisfying $d_{\text{FW}} = p_{\text{W}} - p_{\text{F}}$ and $d_{\text{M}O_{\text{W}}} = \overrightarrow{\text{M} O_{\text{W}} }$, $d_{\text{G}O_{\text{W}}} = \overrightarrow{\text{G} O_{\text{W}} }$.

The overall state vector is $(x,v) \in \real^{28}$ with $x=(p_{\text{W}},~q_{\text{W}},~p_{\text{F}},~q_{\text{F}}) \in \real^{14}$ and $v=(v_{\text{W}},~\dot{q}_{\text{W}},~v_{\text{F}},~\dot{q}_{\text{F}}) = (\dot{p}_{\text{W}},~\dot{q}_{\text{W}},~\dot{p}_{\text{F}},~\dot{q}_{\text{F}})=\dot{x} \in \real^{14}$, where $v_{\text{W}} = \dot{p}_{\text{W}} \in \real^{3}$ represents the linear velocity of the wing in the inertial reference frame, $\dot{q}_{\text{W}} \in \real^{4}$ is the derivative of the quaternion, $q_{\text{W}} \in \real^{4}$ representing the orientation of the wing, $v_{\text{F}} =p_{\text{F}} \in \real^{3}$ is the linear velocity of the fuselage in the inertial reference frame and $\dot{q}_{\text{F}} \in \real^{4}$ is the derivative of the quaternion $q_{\text{F}} \in \real^{4}$ representing the fuselage orientation. It can be seen that the state vector is not minimal. It should be noted that the angular velocity $\omega \in \real^{3}$ can be obtained from the quaternion derivative $\dot{q}$ using equation  \cite[equation (2.7)]{udwadia-schutte} recalled here: 
\begin{align*}
    \omega = H(q) \dot{q} 
\end{align*}
where $H(q) \in \real^{3\times4}$ is a matrix defined by $H(q) = 2\begin{bmatrix}-\epsilon & \eta I_{3} - \skewsym{\epsilon}\end{bmatrix}$.
For deriving the equations of motion, recalling that $R_{i}(q) \in \real^{3}, i = 1,2,3$ are the three columns of a rotation matrix associated with quaternion $q$, define matrices  $L_{\text{i}}^{\text{W}} \left( q_{\text{W}} \right) = \frac{\partial R_{i}}{\partial q}(q_{\text{W}}) \in \real^{3\times4}$, 
$L_{\text{i}}^{\text{F}} \left( q_{\text{F}} \right) = \frac{\partial R_{i}}{\partial q}(q_{\text{F}}) \in \real^{3\times4}$ and
$L_{O_{\text{F}}^{\text{W}}} = \sum_{i=1}^{3} d_{\text{FW}}(i) L_{\text{i}}^{\text{F}} (q_{\text{F}})$, $i \in {1,2,3}$, where $d_{\text{FW}}(i)$ denotes the i-th component of vector $d_{\text{FW}} = p_{\text{W}} - p_{\text{F}}$. Since $O_{\text{W}}$ is located at the wing's center of rotation, the distance $d_{\text{FW}}$ is a constant, since $O_{\text{W}}$ and $O_{\text{F}}$ can be assumed to belong to the same solid (the fuselage). We deduce, with homogeneity, $\dot{L}_{O_{\text{F}}^{\text{W}}} = \sum_{i=1}^{3} d_{\text{FW}}(i) L_{\text{i}}^{\text{F}} (\dot{q}_{\text{F}})$. With these definitions, select the matrices in (\ref{eq:udwadia}) as
\begin{align}
    M = \begin{bmatrix}
        m_{\text{W}} I_{3} & \mathbb{0}_{3 \times 4} & \mathbb{0}_{3} & \mathbb{0}_{3 \times 4}\\
        \mathbb{0}_{4 \times 3} & H_{\text{W}}^\top J_{\text{W}} H_{\text{W}} & \mathbb{0}_{4 \times 3} & \mathbb{0}_{4}\\
        \mathbb{0}_{3} & \mathbb{0}_{3 \times 4} & m_{\text{F}} I_{3} & \mathbb{0}_{3 \times 4} \\
        \mathbb{0}_{4 \times 3} & \mathbb{0}_{4 } & \mathbb{0}_{4 \times 3} & H_{\text{F}}^\top J_{\text{F}} H_{\text{F}}    \end{bmatrix}\in \real^{14\times14},
\end{align}
where we denoted $H_{\text{W}} = H(q_{\text{W}})$, $H_{\text{F}} = H(q_{\text{F}})$, and
\begin{align}
    Q = \begin{bmatrix}
            m_{\text{W}} g e_3 + R(q_{\text{W}}) F_{\text{b}}\\
            -2\dot{H}_{\text{W}}^\top J_{\text{W}} \dot{H}_{\text{W}} \dot{q}_{\text{W}} + H_{\text{W}}^\top M_{\text{W}}\\
            m_{\text{F}} g e_3 + R(q_{\text{F}}) F_{\text{F}}\\
            -2\dot{H}_{\text{F}}^\top J_{\text{F}} \dot{H}_{\text{F}} \dot{q}_{\text{F}} + H_{\text{F}}^\top M_{\text{F}}\\
        \end{bmatrix} \in \real^{14},
\end{align}
% \todo{Details of forces and moments generated while the wing
% transits its position should be given in detail as this is
% the key distinct feature of this aircraft.}
where $\dot{H}_{\text{W}}$ denote $H(\dot{q}_{\text{W}})$, coinciding with the time derivative of $H(q_{\text{W}})$ and $\dot{H}_{\text{F}}$ denote $H(\dot{q}_{\text{F}})$, coinciding with the time derivative of $H(q_{\text{F}})$. Moreover, $F_{\text{b}}$ and $M_{\text{b}}$ represent, respectively,  all the forces and moments acting on the wing. The expressions of $M$ and $Q$ are taken from  \cite[equations (45) and (57)]{doi:10.2514/1.G003374} where the $\phi$ theory is developed, a parametrisation that allows the classical angles of incidence and sideslip to be subtracted and the hover singularity to be avoided. For lack of space, they will not be more detailed. Finally, $F_{\text{F}} =  F_{m}$ et $M_{\text{F}}$ represent respectively the set of non-gravitational forces and moments acting on the fuselage expressed in the frame $O_{\text{W}}$. In particular, $F_{m} = - k_{f} {u_{\text{tail}}}^{2}$ is the force generated by the motor located at the tail of the fuselage and $u_{\text{tail}}$ is the motor rotation speed, while
\begin{align}
    M_{\text{F}} =  m_{\text{F}} g e_3 \times d_{\text{G}O_{\text{W}}} + F_{m} \times d_{\text{M}O_{\text{W}}},
\end{align}
where $d_{\text{M}O_{\text{W}}}$ is the distance between the motor location and the center of rotation and $d_{\text{G}O_{\text{W}}}$ is the distance between the location of the fuselage's center of gravity and the center of rotation.

The set of constraints associated with the nonminimality or the state $(x,v)$ and by the pivot connection between the two bodies is given by:
\begin{align}
    \label{eq:contraintes}
    \left\{
    \begin{aligned}
    &\varphi_{1} \coloneqq q_{\text{W}}^\top q_{\text{W}} - 1 = 0\\
    &\varphi_{2} \coloneqq q_{\text{F}}^\top q_{\text{F}} - 1 = 0\\
    &\varphi_{3} \coloneqq R_{2}(q_{\text{W}})^\top R_{3}(q_{\text{F}}) = 0\\
    &\varphi_{4} \coloneqq R_{2}(q_{\text{W}})^\top R_{1}(q_{\text{F}}) = 0\\
    &\varphi_{5} \coloneqq p_{\text{F}} + d_{\text{FA}} + p_{\text{W}} = 0
    \end{aligned}
    \right.
\end{align}

The first two constraints impose the unit norm of the quaternions $q_{\text{F}}$ and $q_{\text{W}}$.
The third and fourth constraints are related to a moving pivot constraint, i.e. the orthogonality of two vectors is imposed. The last one is a positional constraint so that the point of the centre of rotation belonging to the wing coincides with the point defined in the fuselage. This constraint is based on a three-dimensional geometric closure.\\
It is more convenient to express the set of constraints as a stable dynamical system converging to zero, so we convert each one of the constraints in the form:
\begin{align}
    \ddot{\varphi_{i}} + \delta_{1} \dot{\varphi_{i}}  + \delta_{2} \varphi_{i} = 0, i \in {1,2,3,4,5},
\end{align}
with the selections $(\delta_{1}, \delta_{2}) = (0.5,~8)$ being the coefficients of a stable polynomial, so that, regardless of the selection $\varphi_{i}(0) = 0$, we have $\lim\limits_{t \to \infty} \varphi_{i}(t) = 0$. By differentiating constraints (\ref{eq:contraintes}) twice and factoring them out in the form $ A(x,\dot{x}) \ddot{x} = B(x,\dot{x})$, we obtain the expression of $A(x,\dot{x})$ reported in equation  (\ref{eq:A_contraint}) and $B(x,\dot{x})$ reported in the equation  (\ref{eq:b_contraint}) at the start of the next page.

\begin{align}
\label{eq:A_contraint}
    A = \begin{bmatrix}
            \mathbb{0}_{1 \times 3} & q_{\text{W}}^\top & \mathbb{0}_{1 \times 3} & \mathbb{0}_{1 \times 4}\\
            \mathbb{0}_{1 \times 3} & \mathbb{0}_{1 \times 4} & \mathbb{0}_{1 \times 3} & q_{\text{F}}^\top\\
            \mathbb{0}_{1 \times 3} & R_{3}(q_{\text{F}})^\top L_{2}^{\text{W}}( q_{\text{W}} ) & \mathbb{0}_{1 \times 3} & R_{2}(q_{\text{W}})^\top L_{3}^{\text{F}}( q_{\text{F}} ) \\
            \mathbb{0}_{1 \times 3} & R_{1}(q_{\text{F}})^\top L_{2}^{\text{W}}( q_{\text{W}} ) & \mathbb{0}_{1 \times 3} & R_{2}(q_{\text{W}})^\top L_{3}^{\text{F}}( q_{\text{F}} ) \\
            \mathbb{I}_{3} & L_{O_{\text{F}}^{\text{W}}} & -\mathbb{I}_{3} & \mathbb{0}_{3 \times 4}
        \end{bmatrix}
\end{align}

\begin{figure}[!h]
\centering
    \includegraphics[width=1\columnwidth,angle=0]{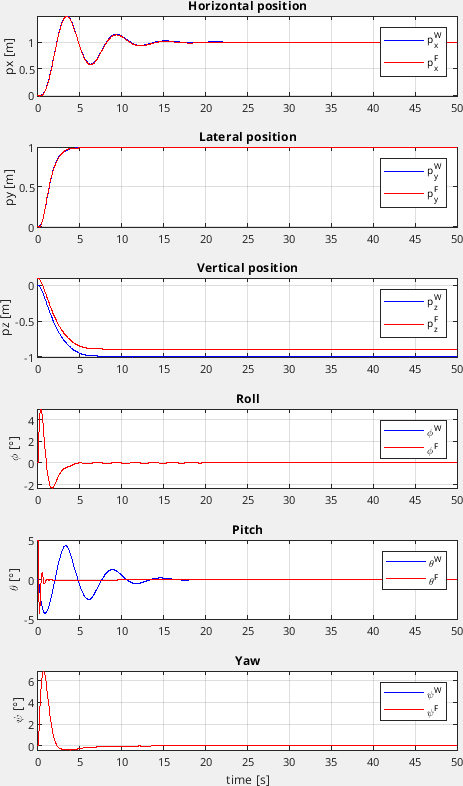}
    \caption{Position and orientation simulation of the multi-body UAV Colibri in closed loop with a simple double-loop controller. }
    \label{fig:sim_colibri}
\end{figure}

The simulation of a drone remains complex, as it is naturally unstable. We have chosen to use the control law proposed in \cite{SANSOU20221} extended to 6 DOF dynamics to stabilize the system. This PI-based control stabilises the wing. Another control law based on a proportional-derivative feedback stabilises the fuselage to keep it horizontal. The closed-loop simulation results are shown in Figure \ref{fig:sim_colibri}.
Considering the degrees of freedom of the pivot link, the coupling between the two bodies is clearly visible from the lower three plots. Indeed, the roll and yaw angles $(\phi_{\text{F}}, \psi_{\text{F}})$ and $(\phi_{\text{W}}, \psi_{\text{W}})$ of the fuselage and wing coincide perfectly, while the pitch angles $(\theta{\text{F}}, \theta{\text{F}})$ are radically different.

\begin{figure*} [!h]
\begin{align}
\label{eq:b_contraint}
    B = \begin{bmatrix}
            -\delta_{1} q_{\text{W}}^\top \dot{q}_{\text{W}} - \frac{\delta_{2}}{2} (q_{\text{W}}^\top q_{\text{W}} -1) - \dot{q}_{\text{W}}^\top \dot{q}_{\text{W}} \\
            -\delta_{1} q_{\text{F}}^\top \dot{q}_{\text{F}} - \frac{\delta_{2}}{2} (q_{\text{F}}^\top q_{\text{F}} -1) - \dot{q}_{\text{F}}^\top \dot{q}_{\text{F}} \\
            -R_{3}(q_{\text{F}})^\top \dot{L}_{2}^{\text{W}}\dot{q}_{\text{W}} - R_{2}(q_{\text{W}})^\top \dot{L}_{3}^{\text{F}} \dot{q}_{\text{F}} - 2\dot{q}_{\text{W}}^\top {L_{2}^{\text{W}}}^\top L_{3}^{\text{F}} \dot{q}_{\text{F}} - \delta_{1}(R_{3}(q_{\text{F}})^\top L_{2}^{\text{W}}\dot{q}_{\text{W}} +  R_{2}(q_{\text{W}})^\top L_{3}^{\text{F}} \dot{q}_{\text{F}} ) - \delta_{2}\varphi_{3}\\
            -R_{1}(q_{\text{F}})^\top \dot{L}_{2}^{\text{W}}\dot{q}_{\text{W}} - R_{2}(q_{\text{W}})^\top \dot{L}_{1}^{\text{F}} \dot{q}_{\text{F}} - 2\dot{q}_{\text{W}}^\top {L_{2}^{\text{W}}}^\top L_{1}^{\text{F}} \dot{q}_{\text{F}} - \delta_{1}(R_{1}(q_{\text{F}})^\top L_{2}^{\text{W}}\dot{q}_{\text{W}} +  R_{2}(q_{\text{W}})^\top L_{1}^{\text{F}} \dot{q}_{\text{F}} ) - \delta_{2} \varphi_{4}\\
            \dot{L}_{O_{\text{F}}^{\text{W}}} \dot{q}_{\text{F}}  - \delta_{1}( v_{\text{W}} + \dot{L}_{O_{\text{F}}^{\text{W}}} \dot{q}_{\text{W}} - v_{\text{F}}) - \delta_{1}\varphi_{5}
        \end{bmatrix}
\end{align}
  \hrulefill\par
\end{figure*}

\section{State estimation}\label{sec:stateEst}
% \todo{Link to the modeling:  What is the connection between angular velocity
% estimation and the multibody dynamics derived in the paper?
% What are the advantages of the angular velocity estimator
% compared to existing technologies? It is difficult to
% discern the theoretical contribution to the angular
% velocity estimation part.}
In order to stabilise this two-body UAV system, it is necessary to know the position and orientation of the two bodies. Due to the pivot link between the wing and the fuselage, the difference between the orientation of the wing and the orientation of the fuselage is simply a rotation about the pitch axis of the wing. The two other orientations (roll and yaw) coincide. The position of the fuselage's centre of gravity can be deduced from the position of the wing's centre of gravity and the angle between the fuselage and the wing. This angle is measured by a quadrature rotary encoder (CUI Devices AMT22, Absolute Encoders, 12 bit, SPI), which returns a quantized angular measurement with a step size of \SI{0.09}{\degree}. Given this angular measurement, we discuss below the estimation of the speed information, so as to reconstruct the state of the UAV. 

\subsection{Sensors placement}
\label{subsec:sens_pos}
A first question pertains to the sensors placements: the IMU (accelerometer, gyroscope and magnetometer) can be installed on the fuselage or on the wing. 
Installing the IMU on the wing means that the measurements can be taken directly in the desired reference frame, but the measurements are noisier because the IMU is attached to the structure supporting the motors. Given the size of the wing, their flexibility can generate resonances and can perturb the measurements. 
%In addition, wiring the IMU is not easy as the cables must pass through the pivot link, which generates resistive torques.
Installing the IMU on the fuselage reduces vibrations, but means that the measurements must be transformed in the wing reference frame. The corresponding transformation can be computed from the rotary encoder measurement, providing the angle between the wing and the fuselage, and also from the measurements taken with the CAD software, providing precise information about the distances between the wing and fuselage frames. Our final choice is to attach the IMU to the fuselage. Another consideration is that the autopilot board, which already have an integrated IMU, is also supposed to be connected to the payload and other sensors attached to the fuselage. It is thus limiting the number of cables at the pivot point to the actuators commands and power supply.

\subsection{Angular speed estimation}
\begin{figure}[h]
\centering
    \includegraphics[width=1\columnwidth,angle=0]{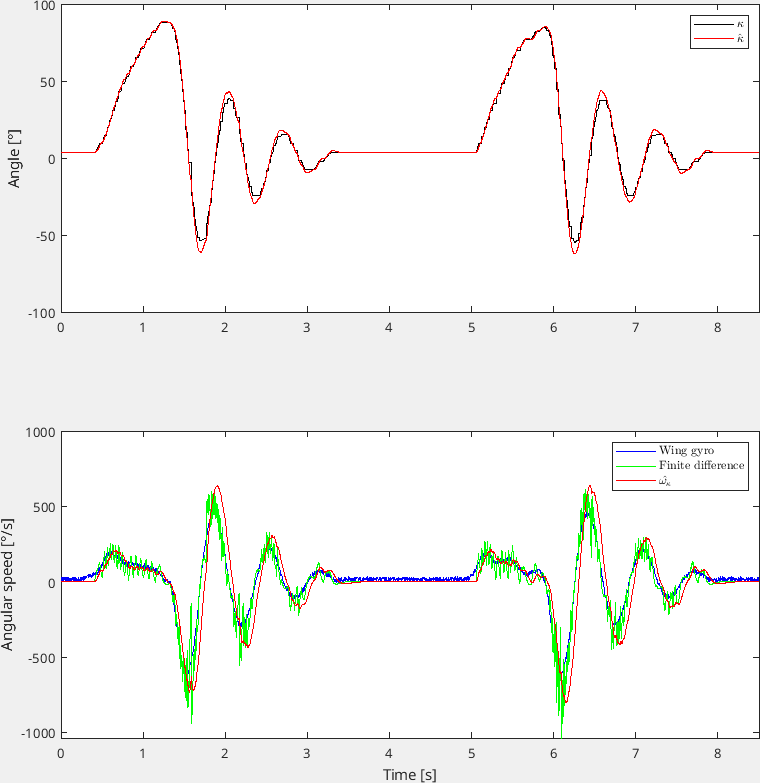}
    \caption{Angular position measurement (black,top plot), wing gyro velocity measurement (blue,bottom plot), finite difference velocity estimation (green, bottom plot) and high-gain estimates (red curves)}
    \label{fig:high_gain}
\end{figure}

As explained above, we can measure the angle $\kappa \in \real$ between the wing and the fuselage using the rotary encoder. Then, to estimate the angular velocity we use the high-gain observer proposed in \cite{203613} (see also \cite{1032320} for the use of high-gain observers to estimate time derivatives). This method is preferable to a finite difference derivative, as the quantized information generated by the rotary encoder can result in bursts in the estimated angular velocity values.\\ 
Denote by $\kappa \in \real$ the measured position variable, by $\omega_{\kappa} := \dot \kappa  \in \real$ its derivative, to be estimated, and by $\xi = [\kappa,~\omega_{\kappa}]^\top \in \mathbb{R}^2$ their juxtaposition in a single vector. Denote also $\hat{\xi}$ the estimate of $\xi$ as follows:
%\todo{GH: il y a pas un peu de répétitions dans les notations ?}
%\xi = [\kappa,~\omega_{\kappa}]^\top \in \mathbb{R}^2, \quad 
\begin{align*}
    \hat{\xi} = [\hat{\kappa},~\hat{\omega}_{\kappa}]^\top \in \mathbb{R}^2.
\end{align*}
Following \cite{203613}, the estimator dynamics is given by
\begin{align}
\label{eq:high_dyn}
    \dot{\hat{\xi}} =  \begin{bmatrix}0 & 1 \\ 0 & 0 \end{bmatrix} \hat{\xi}+ \begin{bmatrix}\frac{k_{p}}{\epsilon_{\kappa}}  \\ \frac{k_{v}}{\epsilon_{\kappa}^{2}}  \end{bmatrix} (\kappa - \hat{\kappa}),
\end{align}
where $\kappa$ is the angular measurement recovering from the sensors, $k_{p}$ and $k_{v}$ are two positive scalars gains such that the characteristic equation $s^{2} + k_{v} s + k_{p} = 0$ has roots with negative real part. For our estimators, we have selected $k_{p} = 1$ and $k_{v} = 1.3$ so as to get a damping factor $\zeta = 0.65$ leading to a slightly underdamped response as a suitable trade-off between a fast rise time and a mildly oscillatory response. The high-gain scaling factor
$\epsilon_{\kappa}$ can be conveniently adjusted in order to obtain a trade-off between smoothing action (obtained by increasing $\epsilon_{\kappa}$) and reduction of the time lag
of the estimator (obtained by reducing $\epsilon_{\kappa}$). Moreover, the smoothing action of the proposed approach mitigates the effect of the quantized position measurements. We have selected $\epsilon_{\kappa} = 0.05$ for our experiments. Figure \ref{fig:high_gain} shows the experimental results obtained after implementation of the high-gain filter (\ref{eq:high_dyn}) in the case of a flight generating high-amplitude angular oscillations.
%\todo{GH: c'est pas très lisible comme figure}
We carried out differentiation by finite difference (in green) in post-treatment to compare the results. Due to the quantized nature of the rotary encoder, we observe that the angular velocity obtained by finite difference is very noisy. We can see that the high-gain filter makes it possible to estimate the angular velocity more accurately (in red), albeit with a slight delay. Thanks to the addition of an extra IMU on the wing in a specific flight test, it is possible to compare the velocity estimate with the wing's gyroscope (MPU9250) measurements, visible on the bottom graph of Figure \ref{fig:high_gain} (blue trace). We can see that the gyroscope readings are somewhat noisy, due in particular to the vibrations generated by the motors.  %\todo{GH: ref à la figure est trop tard par rapport au texte}
\\
In order to perform the necessary transformation among the reference frames, define the quaternion $q_{\hat{\kappa}} \in {\mathbb S}^3$ as follows:
\begin{align}
\label{eq:rot_quat}
    q_{\hat{\kappa}} =  \left [\cos\left(\frac{\hat{\kappa}}{2}\right) ~ 0 ~ \sin\left(\frac{\hat{\kappa}}{2}\right) ~ 0 \right]^\top
\end{align}

\subsection{Wing state estimation}
Based on the estimated angle $\hat{\kappa}$ and the estimated angular velocity $\hat{\omega}_{\kappa}$, it is possible to transform the measurements from the fuselage to the wing frame. All the sensors are installed on the autopilot board, which is itself attached to the fuselage. However, as mention in introduction, we want to use INDI to stabilize the wing. So this control law requires the state information in the wing reference frame, where all the forces are applied (aerodynamic and traction).
Then, two viable solution are possible: perform the state estimation in the fuselage reference frame and rotate the estimation, using the estimate of the angle $\hat{\kappa}$, or rotate the raw measurements in advance to express them in the wing reference frame, and then perform the state estimation on the latter. 
Given the current architecture of the software in the Paparazzi\footnote{\url{https://github.com/enacuavlab/paparazzi/tree/rot_state_est}} system, it is cumbersome to have two joint state estimation structures, so it is difficult to implement the first solution, where the controller directly retrieves the current state estimation. For this reason, we have chosen to estimate the state of the wing from data measured on the fuselage. To this end, we detail below the coordinate transformation for the three sensors: gyroscope, accelerometer and magnetometer. \\
\indent For the gyroscope-based angular rate measurements, we may compute the angular velocity of the wing expressed in the wing frame as
\begin{align}
    \label{eq:gyro_deplacement}
    \omega_{\text{W}} = R(q_{\hat{\kappa}}) \left( \omega_{gyro}^{\text{F}} + \begin{bmatrix}
    0\\ \omega_{\kappa} \\ 0
    \end{bmatrix}  \right) 
\end{align}
where $\omega_{gyro}^{\text{F}}$ is the angular velocity measured by the gyro on the fuselage, expressed in the fuselage frame, $\hat{\omega}_{\kappa}$ is the estimated angular velocity of the wing relative to the fuselage, as per (\ref{eq:high_dyn}), and $q_{\hat{\kappa}}$ is the quaternion defined in (\ref{eq:rot_quat}).
Expression (\ref{eq:gyro_deplacement}) is similar to a composition of angular velocities and a reference frame transformation.\\
\indent For the acceleration measurement with the accelerometer, we may use the following relation
Expression (\ref{eq:accel_deplacement}) is obtained from the rate of change transport theorem \cite{brizard2004motion}, where we find the Euler acceleration term $\dot{\omega_{\text{F}}} \times d_{AF}$ and the centripetal acceleration term $\omega_{\text{F}} \times ( \omega_{\text{F}} \times  d_{AF})$. Coriolis Acceleration $2\omega_{\text{F}} \times \frac{d (d_{\text{FW}})}{d t}\Bigr|_{O_{\text{F}}}$ and the rate of acceleration $\frac{d^{2} (d_{\text{FW}})}{d^{2} t}\Bigr|_{O_{\text{F}}}$ are zero because $d_{\text{FW}}$ is contant.
\begin{align}
    \label{eq:accel_deplacement}
    a_{\text{W}} = R(q_{\hat{\kappa}}) \left( a_{acc}^{F} + \dot{\omega}_{gyro}^{F} \times d_{\text{FW}} + \omega_{gyro}^{F} \times ( \omega_{gyro}^{F} \times  d_{\text{FW}}) \right) 
\end{align}
where $a_{acc}^{F} \in \real^{3}$ is the acceleration measured by the accelerometer on the fuselage, expressed in the fuselage frame and $\omega_{gyro}^{F}$, the angular velocity of the fuselage, same as the equation (\ref{eq:gyro_deplacement}). The angular acceleration $\dot{\omega}_{gyro}^{F}$ in (\ref{eq:accel_deplacement}) is computed by a finite difference.\\
\indent For the magnetometer measurements, we have
\begin{align}
    \label{eq:mag_deplacement}
    E_{\text{W}} = R(q_{\hat{\kappa}}) E_{mag}
\end{align}
where $E_{mag} \in \real^{3}$ is the magnetometer output, expressed in the fuselage frame and  $ E_{\text{W}} \in \real^{3}$ is the computed measurement expressed in the wing frame.\\
\indent To obtain the wing state estimate, we use a sensor measurement fusion algorithm: extended Kalman filter\footnote{\url{https://github.com/PX4/PX4-ECL/tree/master}} (EKF) which provide an estimate of the following states: $p_{\text{W}}$, $v_{\text{W}}$, $q_{\text{W}}$ from measurements transformed in the wing reference frame $\omega_{\text{W}}$ (eq. (\ref{eq:gyro_deplacement})), $a_{\text{W}}$ (eq. (\ref{eq:accel_deplacement})), $E_{\text{W}}$ (eq. (\ref{eq:mag_deplacement})) and external vision system pose data, which provides a precise measurement of the drone's position $p_{\text{W}}$ and speed $v_{\text{W}}$ in the inertial reference frame (I).

\subsection{Fuselage orientation estimation}
To determine the orientation of the fuselage, we may perform a composition between the quaternion representing the orientation of the wing $q_{\text{W}}$ result of EKF and the quaternion constructed from the filtered measurement of the rotary encoder $q_{\hat{\kappa}}$ in (\ref{eq:rot_quat}),
\begin{align}
\label{eq:quat_fuselage}
    q_{\text{F}} = q_{\text{W}} \otimes q_{\hat{\kappa}}
\end{align}
where the operator $\otimes$ denotes the qaternion product. The knowledge of $q_{\text{F}}$ is needed to keep the fuselage perfectly horizontal. 

\section{Incremental Nonlinear Dynamic Inversion combined with PD pendulum control}
\label{sec:control}

The theory of Incremental Nonlinear Dynamic Inversion (INDI) used in the context of micro-UAVs is presented in \cite{doi:10.2514/1.G001490}. We use the notation proposed in \cite{doi:10.2514/1.G004520}, without providing extra details, due to length constraints. The central underlying assumption is that the so-called timescale separation principle holds w.r.t. the actuator dynamics and the dynamics of aerodynamic forces and moments. The control signal can then be computed incrementally using the actuator effectiveness matrix $G$.
\begin{align}
    u_{\text{W}} = u_{\text{W}} + G^{\dag} (\nu - \begin{bmatrix}
    \dot{\omega}_{\text{W}} \\
    T_{\text{W}}
    \end{bmatrix})
\end{align}
where $ \dot{\omega}_{\text{W}} \in \real^{3}$ is the measured angular acceleration obtain by finite difference from equation (\ref{eq:gyro_deplacement}), $T_{\text{W}} \in \real$ is the current thrust, $\nu$ is define in \cite[equation (4)]{doi:10.2514/1.G004520} and $G$ is the control effectiveness matrix, determined as follows :
\begin{align*}
    \begin{bmatrix}
    \partial \phi \\
    \partial \theta \\
    \partial \psi \\
    \partial T
    \end{bmatrix}\! =\! G u_{f} \!=\!
    \begin{bmatrix}
    -7.5 & -15 & 7.5 & 15 & 0 & 0\\
    0 & 0 & 0 & 0 & 15 & 15 \\
    0 & 0 & 0 & 0 & 4 & -4 \\
    -0.6 & -0.6 & -0.6 & -0.6 & 0 & 0\\
    \end{bmatrix}
    u_{f}
\end{align*}
This selection of efficiency matrix has been determined for the hovering flights, but it is necessary to carry out a different study for the forward flight.

To stabilise the fuselage, we use a PD feedback from the angle $\theta_{\text{F}}$ formed between the fuselage and the horizontal, which we want to keep at zero. This is obtained by converting the quaternion $q_{\text{F}}$ of equation (\ref{eq:quat_fuselage}) into an Euler angle by following the 'ZYX' Euler convention. The PD feedback provides the reference $u_{\text{tail}}$ for the angular speed of the motor generating the force $F_{m}$ (see Figure  \ref{fig:colibri_frame_side}), as follows
\begin{align*}
    u_{\text{tail}} = u_{eq} + k_{p} \theta_{\text{F}} + k_{d} \dot{\theta_{\text{F}}},
\end{align*}
where $u_{eq}$ is the equilibrium motor command to keep the fuselage horizontal in the absence of disturbance and $k_{p}$, $k_{d}$ are tunable scalar gains. The value $u_{eq}$ was obtained by applying a moment theorem to the fuselage at the point $O_{\text{W}}$. In fact, the two moments that come into effect on the fuselage are the torque due to the thrust force of the tail motor and the torque due to the position of the fuselage's centre of gravity.
The gains $k_{p}$ et $k_{d}$ were adjusted in flight to ensure satisfactory flight behaviour. We obtain $\dot{\theta_{\text{F}}}$ from $\omega_{gyro}^{F} = [\dot{\phi_{\text{F}}}~\dot{\theta_{\text{F}}}~\dot{\psi_{\text{F}}}]^\top$.
\section{Experimentation}
\label{sec:exp}
\begin{figure}[!h]
    \centering
    \includegraphics[trim={0 15cm 0 25cm},clip, width=\columnwidth]{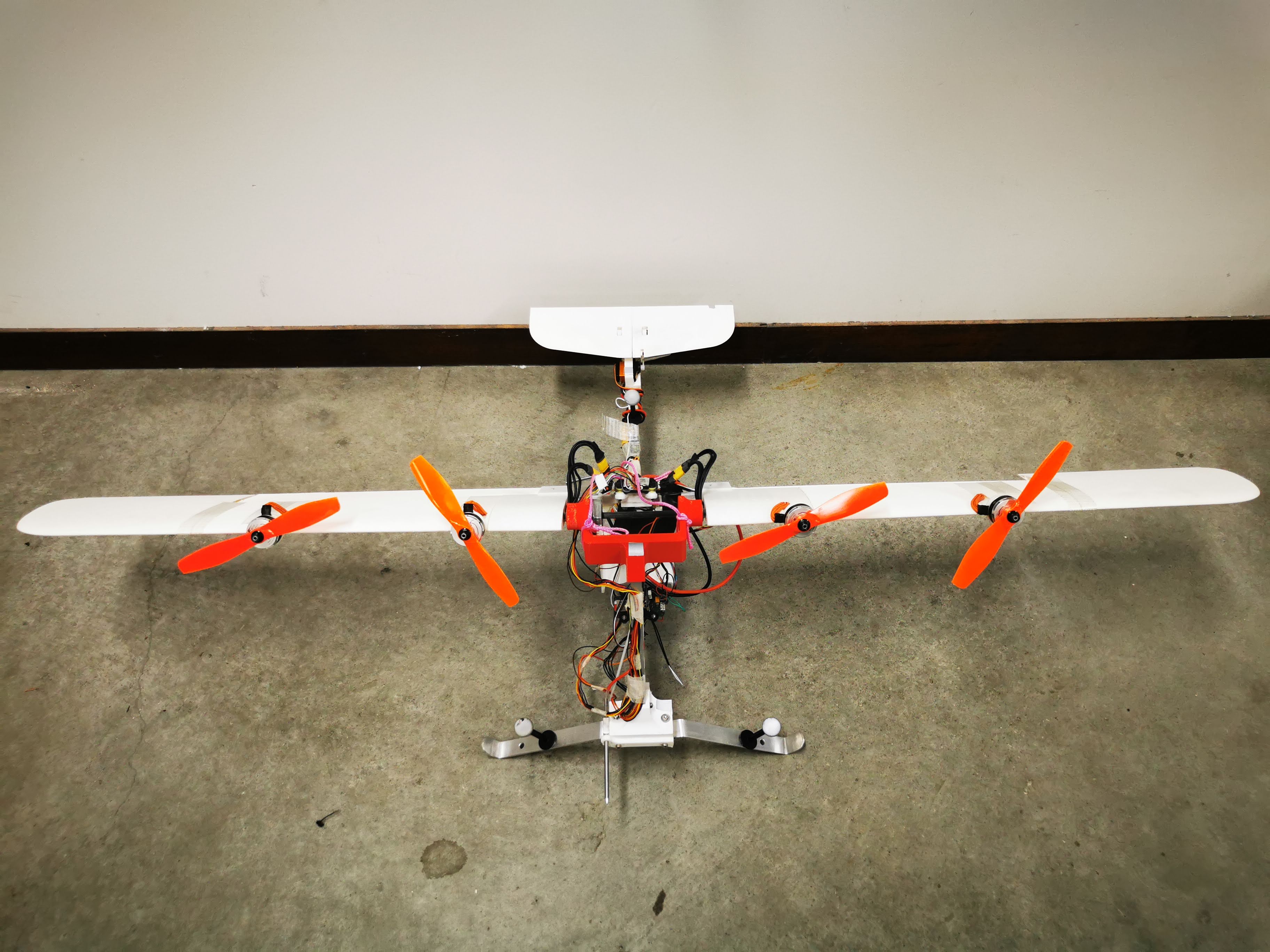}
    \caption{Colibri experimental prototype.}
    \label{fig:colibri_real}
\end{figure}

\begin{figure}[h]
\centering
    \includegraphics[width=1\columnwidth,angle=0]{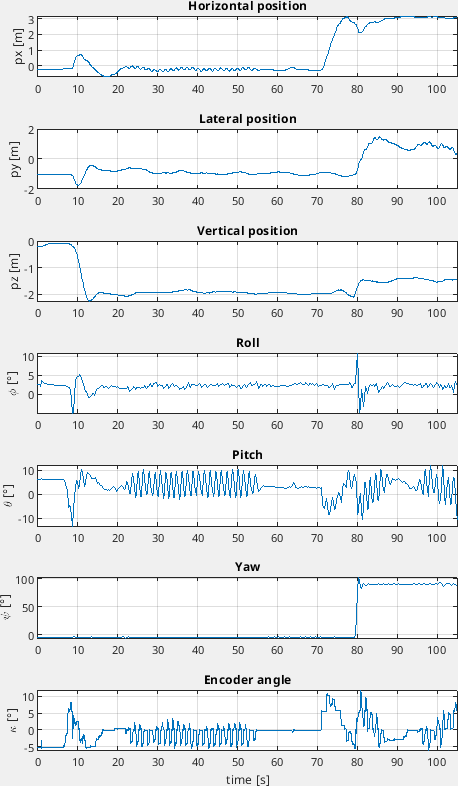}
    \caption{Position and orientation of the reference frame wing in the first six graphs and pivot angle measurement on the last graph below during real flight. }
    \label{fig:colibri_flight}
\end{figure}

% \todo{Photo de la volière: It is difficult to understand the experimental results
% due to the absence of snapshots, detailed explanations, or
% pictures of the experimental environment. Illustrations and
% explanations of the experimental structure, actual flight
% photos, and the design structure of the entire flight
% system must be added.}
An experimental prototype was developed, as shown in Figure~\ref{fig:colibri_real}. A selection of the experimental results in controlled flight is shown in Figure \ref{fig:colibri_flight}.\\
About Figure \ref{fig:colibri_flight}, from \SI{0}{\second} to \SI{8}{\second}, the drone is on the ground. From \SI{8}{\second} to \SI{16}{\second}, the drone takes off to reach a height of 2 metres visible from the third plot. This height is reached after a 10 \% overshoot. The drone is held in this position for \SI{54}{\second}. Incidence oscillations are observed in the fifth and last plot, generating oscillations in the drone's horizontal position. This is due to the coupling between the two bodies, which is not properly stabilized. From \SI{70}{\second}, the UAV starts heading towards the point $p_{c} = \begin{bmatrix} 3 & 0.9 & -1.5 \end{bmatrix}^\top$ and $\psi_{c}=\SI{90}{\degree}$.
%This brings the drone in front of the WindShape open wind tunnel. The wind generator is switched on at \SI{95}{\second} at a speed of \SI{1}{\per\second}. This disturbance destabilises the drone, which is not shown here. \todo{GH: ça fait beaucoup d'info qui ne seront pas détaillés, est-ce qu'il faut tout garder ?}

\section{Conclusions and Future Work}\label{conclusion}
We present a novel architecture of a freewing UAV with a powered fuselage, with the aim of keeping the fuselage as stable as possible. Modelling based on the Udwadia-Kalaba equations has been used to describe the UAV multi-body system dynamics. We plan to use this model with the addition of constraints to obtain control law that will cover the entire flight domain.
We also described the state estimation method used for this multi-body UAV, with a comparison between ground truth measurements temporarily placed on the wing and on the fuselage. 
The experimental phase enabled validating the hovering behaviour of the UAV without disturbances using two control loops. However, we observed the limits of the double loop architecture in the presence of disturbances, such as unmodeled forward wind generated by an open wind tunnel. 
Future work includes designing a centralized control architecture that will enable us to control the UAV over its entire flight domain, by taking into account the coupling between the two bodies.
% \todo{Where is the disturbance introduced in
% tha paper? What is the interconnection between the tail
% control input and the wing control input? What do you mean
% by decentralized in the proposed approach?}

\section*{ACKNOWLEDGMENT}
The authors would like to thank Murat Bronz for useful discussions.
%\todo{mettre à la fin}


%%%%%%%%%%%%%%%%%%%%%%%%%%%%%%%%%%%%%%%%%%%%%%%%%%%%%%%%%%%%%%%%%%%%%%%%%%%%%%%%

%%%%%%%%%%%%%%%%%%%%%%%%%%%%%%%%%%%%%%%%%%%%%%%%%%%%%%%%%%%%%%%%%%%%%%%%%%%%%%%%

%%%%%%%%%%%%%%%%%%%%%%%%%%%%%%%%%%%%%%%%%%%%%%%%%%%%%%%%%%%%%%%%%%%%%%%%%%%%%%%%


%%%%%%%%%%%%%%%%%%%%%%%%%%%%%%%%%%%%%%%%%%%%%%%%%%%%%%%%%%%%%%%%%%%%%%%%%%%%%%%%

\bibliographystyle{IEEEtran}
\bibliography{mybibfile}

\begin{thebibliography}{10}
\providecommand{\url}[1]{#1}
\csname url@samestyle\endcsname
\providecommand{\newblock}{\relax}
\providecommand{\bibinfo}[2]{#2}
\providecommand{\BIBentrySTDinterwordspacing}{\spaceskip=0pt\relax}
\providecommand{\BIBentryALTinterwordstretchfactor}{4}
\providecommand{\BIBentryALTinterwordspacing}{\spaceskip=\fontdimen2\font plus
\BIBentryALTinterwordstretchfactor\fontdimen3\font minus
  \fontdimen4\font\relax}
\providecommand{\BIBforeignlanguage}[2]{{%
\expandafter\ifx\csname l@#1\endcsname\relax
\typeout{** WARNING: IEEEtran.bst: No hyphenation pattern has been}%
\typeout{** loaded for the language `#1'. Using the pattern for}%
\typeout{** the default language instead.}%
\else
\language=\csname l@#1\endcsname
\fi
#2}}
\providecommand{\BIBdecl}{\relax}
\BIBdecl

\bibitem{patentZuck}
Z.~D. R, ``Airplane with nonstalling and glide angle control characteristics,''
  U.S. Patent 2347230A, Apr. 1944.

\bibitem{Date1972ExperimentalIO}
\BIBentryALTinterwordspacing
R.~Date, R.~A. Ormiston, B.~M.~O. Name, U.~S. Army, A.~Mobility, D.~Laboratory,
  M.~Field, D.~C. Washington, .-G. Classif, and U.~Unclassified, ``Experimental
  investigation of stability and stall flutter of a free-floating wing v/stol
  model,'' 1972. [Online]. Available:
  \url{https://api.semanticscholar.org/CorpusID:117259271}
\BIBentrySTDinterwordspacing

\bibitem{Delair_2024}
\BIBentryALTinterwordspacing
Feb 2024. [Online]. Available:
  \url{https://delair.aero/delair-commercial-drones/dt46-long-range-made-easy/}
\BIBentrySTDinterwordspacing

\bibitem{doi:10.2514/6.1992-4194}
\BIBentryALTinterwordspacing
W.~CHEN and J.~BARLOW, \emph{An ultralight freewing aircraft design study}.
  [Online]. Available: \url{https://arc.aiaa.org/doi/abs/10.2514/6.1992-4194}
\BIBentrySTDinterwordspacing

\bibitem{doi:10.2514/6.2005-1024}
\BIBentryALTinterwordspacing
K.~Ro and J.~Barlow, \emph{Dynamic Modeling of Articulated Wing-Body-Tail
  Unmanned Aerial Vehicle}. [Online]. Available:
  \url{https://arc.aiaa.org/doi/abs/10.2514/6.2005-1024}
\BIBentrySTDinterwordspacing

\bibitem{RoKammanJames}
K.~Ro, J.~Kamman, and J.~Barlow, ``Flight perfornance analysis of freewing
  tilt-body unmanned aerial vehicle,'' 08 2005.

\bibitem{Leylek2015UseOC}
\BIBentryALTinterwordspacing
E.~A. Leylek and M.~Costello, ``Use of compliant hinges to tailor flight
  dynamics of unmanned aircraft,'' \emph{Journal of Aircraft}, vol.~52, pp.
  1692--1706, 2015. [Online]. Available:
  \url{https://api.semanticscholar.org/CorpusID:46635008}
\BIBentrySTDinterwordspacing

\bibitem{HavBerJoh}
S.~Haviland, D.~Bershadsky, and E.~Johnson, ``Dynamic modeling and analysis of
  a vtol freewing concept,'' 01 2016.

\bibitem{doi:10.2514/6.2020-1263}
\BIBentryALTinterwordspacing
R.~M. Axten and E.~N. Johnson, \emph{Stabilizing a VTOL Freewing Testbed
  Vehicle in Hover}. [Online]. Available:
  \url{https://arc.aiaa.org/doi/abs/10.2514/6.2020-1263}
\BIBentrySTDinterwordspacing

\bibitem{doi:10.2514/6.2024-0726}
\BIBentryALTinterwordspacing
R.~M. Axten, T.~Khamvilai, and E.~Johnson, \emph{Multi-Outer Loop Adaptive
  Control for a VTOL Free-Wing Aircraft}. [Online]. Available:
  \url{https://arc.aiaa.org/doi/abs/10.2514/6.2024-0726}
\BIBentrySTDinterwordspacing

\bibitem{McSwain2017GreasedL}
\BIBentryALTinterwordspacing
R.~G. McSwain, L.~J. Glaab, C.~R. Theodore, R.~D. Rhew, and D.~D. North,
  ``Greased lightning (gl-10) performance flight research: Flight data
  report,'' 2017. [Online]. Available:
  \url{https://api.semanticscholar.org/CorpusID:115523083}
\BIBentrySTDinterwordspacing

\bibitem{binz2019attitude}
F.~Binz, T.~Islam, and D.~Moormann, ``Attitude control of tiltwing aircraft
  using a wing-fixed coordinate system and incremental nonlinear dynamic
  inversion,'' \emph{International Journal of Micro Air Vehicles}, vol.~11, p.
  1756829319861370, 2019.

\bibitem{akr2022DesignAA}
\BIBentryALTinterwordspacing
H.~Çakır and D.~F. Kurtuluş, ``Design and aerodynamic analysis of a vtol
  tilt-wing uav,'' \emph{Turkish J. Electr. Eng. Comput. Sci.}, vol.~30, pp.
  767--784, 2022. [Online]. Available:
  \url{https://api.semanticscholar.org/CorpusID:239845307}
\BIBentrySTDinterwordspacing

\bibitem{articleSimmonsMurphy}
B.~Simmons and P.~Murphy, ``Aero-propulsive modeling for tilt-wing, distributed
  propulsion aircraft using wind tunnel data,'' \emph{Journal of Aircraft},
  vol.~59, pp. 1--17, 03 2022.

\bibitem{schutt2014fullscale}
M.~Sch{\"u}tt, P.~Hartmann, and D.~Moormann, ``Fullscale windtunnel
  investigation of actuator effectiveness during stationary flight within the
  entire flight envelope of a tiltwing mav,'' in \emph{International Micro Air
  Vehicle Competition and Conference}, 2014, pp. 77--83.

\bibitem{udwadia-phohomsiri}
\BIBentryALTinterwordspacing
F.~E. Udwadia and P.~Phohomsiri, ``Explicit equations of motion for constrained
  mechanical systems with singular mass matrices and applications to multi-body
  dynamics,'' \emph{Proceedings: Mathematical, Physical and Engineering
  Sciences}, vol. 462, no. 2071, pp. 2097--2117, 2006. [Online]. Available:
  \url{http://www.jstor.org/stable/20208995}
\BIBentrySTDinterwordspacing

\bibitem{udwadia-schutte}
\BIBentryALTinterwordspacing
F.~E. Udwadia and A.~D. Schutte, ``A unified approach to rigid body rotational
  dynamics and control,'' \emph{Proceedings: Mathematical, Physical and
  Engineering Sciences}, vol. 468, no. 2138, pp. 395--414, 2012. [Online].
  Available: \url{http://www.jstor.org/stable/41345881}
\BIBentrySTDinterwordspacing

\bibitem{udwadia-koganti}
\BIBentryALTinterwordspacing
P.~B. Koganti and F.~E. Udwadia, ``Dynamics and precision control of uncertain
  tumbling multibody systems,'' \emph{Journal of Guidance, Control, and
  Dynamics}, vol.~40, no.~5, pp. 1176--1190, 2017. [Online]. Available:
  \url{https://doi.org/10.2514/1.G002212}
\BIBentrySTDinterwordspacing

\bibitem{doi:10.2514/1.G003374}
\BIBentryALTinterwordspacing
L.~R. Lustosa, F.~Defa\"{y}, and J.-M. Moschetta, ``Global singularity-free
  aerodynamic model for algorithmic flight control of tail sitters,''
  \emph{Journal of Guidance, Control, and Dynamics}, vol.~42, no.~2, pp.
  303--316, 2019. [Online]. Available: \url{https://doi.org/10.2514/1.G003374}
\BIBentrySTDinterwordspacing

\bibitem{SANSOU20221}
\BIBentryALTinterwordspacing
F.~Sansou, F.~Demourant, G.~Hattenberger, T.~Loquen, and L.~Zaccarian, ``Open
  wind tunnel experiments of the darko tail-sitter longitudinal stabilization
  with constant wind,'' \emph{IFAC-PapersOnLine}, vol.~55, no.~22, pp. 1--6,
  2022, 22nd IFAC Symposium on Automatic Control in Aerospace ACA 2022.
  [Online]. Available:
  \url{https://www.sciencedirect.com/science/article/pii/S2405896323002598}
\BIBentrySTDinterwordspacing

\bibitem{203613}
S.~Nicosia, A.~Tornambe, and P.~Valigi, ``Experimental results in state
  estimation of industrial robots,'' in \emph{29th IEEE Conference on Decision
  and Control}, 1990, pp. 360--365 vol.1.

\bibitem{1032320}
Y.~Chitour, ``Time-varying high-gain observers for numerical differentiation,''
  \emph{IEEE Transactions on Automatic Control}, vol.~47, no.~9, pp.
  1565--1569, 2002.

\bibitem{brizard2004motion}
A.~J. Brizard, ``Motion in a non-inertial frame,'' \emph{Saint Michael's
  College, Colchester, VT}, 2004.

\bibitem{doi:10.2514/1.G001490}
\BIBentryALTinterwordspacing
E.~J.~J. Smeur, Q.~Chu, and G.~C. H.~E. de~Croon, ``Adaptive incremental
  nonlinear dynamic inversion for attitude control of micro air vehicles,''
  \emph{Journal of Guidance, Control, and Dynamics}, vol.~39, no.~3, pp.
  450--461, 2016. [Online]. Available: \url{https://doi.org/10.2514/1.G001490}
\BIBentrySTDinterwordspacing

\bibitem{doi:10.2514/1.G004520}
\BIBentryALTinterwordspacing
E.~J.~J. Smeur, M.~Bronz, and G.~C. H.~E. de~Croon, ``Incremental control and
  guidance of hybrid aircraft applied to a tailsitter unmanned air vehicle,''
  \emph{Journal of Guidance, Control, and Dynamics}, vol.~43, no.~2, pp.
  274--287, 2020. [Online]. Available: \url{https://doi.org/10.2514/1.G004520}
\BIBentrySTDinterwordspacing

\end{thebibliography}

\end{document}